\documentclass{article}

\PassOptionsToPackage{numbers, compress}{natbib}

\PassOptionsToPackage{hyperref}{breaklinks}
\usepackage[preprint]{neurips_2024}

\usepackage[utf8]{inputenc} 
\usepackage[T1]{fontenc}    
\usepackage{hyperref}       
\usepackage{url}            
\usepackage{booktabs}       
\usepackage{amsfonts}       
\usepackage{nicefrac}       
\usepackage{microtype}      
\usepackage{xcolor}         
\usepackage{amsmath}
\usepackage{xspace}
\usepackage[most]{tcolorbox}
\usepackage{soul}
\usepackage{array}
\usepackage{placeins}

\newtcolorbox{white}{colback=white!10!white,boxrule=0pt, top=0pt,bottom=0pt, left=0pt}
\newtcolorbox{blue}{colback=blue!10!white,boxrule=0pt, top=0pt,bottom=0pt, left=0pt}
\newtcolorbox{red}{colback=red!10!white,boxrule=0pt,top=0pt,bottom=0pt, left=0pt}
\newtcolorbox{green}{colback=green!10!white,boxrule=0pt, top=0pt,bottom=0pt, left=0pt}

\colorlet{bluec}{blue!20!white}
\colorlet{greenc}{green!20!white}
\colorlet{redc}{red!20!white}

\newcommand\ba{\mathcal{B}}
\newcommand\ke{\mathcal{K}}

\newcommand{\wrld}{\textit{WorldLM}\xspace}

\usepackage{setspace}

\usepackage{algorithm}

\usepackage[noend]{algpseudocode}

\algdef{SE}[SUBALG]{Indent}{EndIndent}{}{\algorithmicend\ }%
\algtext*{Indent}
\algtext*{EndIndent}

\usepackage[capitalise]{cleveref} 
\usepackage{graphicx}
\usepackage{subcaption}
\usepackage{multirow}

\usepackage{siunitx}

\title{Worldwide Federated Training of Language Models}

\makeatletter
\newcommand{\myfnsymbol}[1]{%
  \expandafter\@myfnsymbol\csname c@#1\endcsname
}

\newcommand{\@myfnsymbol}[1]{%
  \ifcase #1
  \or 1
  \or 2
  \or \TextOrMath{\textasteriskcentered}{*}
  \or \TextOrMath{\textasteriskcentered}{*}\TextOrMath{\textasteriskcentered}{*}
  \or \TextOrMath{\textdagger}{\dagger}
  \or \TextOrMath{\textasteriskcentered}{*},\TextOrMath{\textasteriskcentered}{*}\TextOrMath{\textasteriskcentered}{*}
  \fi
}

\newcommand{\affiliationA}{\@myfnsymbol{1}}
\newcommand{\affiliationB}{\@myfnsymbol{2}}
\newcommand{\equalcontributor}{\@myfnsymbol{3}}
\newcommand{\biequalcontributor}{\@myfnsymbol{4}}
\newcommand{\correspondingA}{\@myfnsymbol{5}}

\author{
Alex Iacob\textsuperscript{\affiliationA,\affiliationB}
\And 
Lorenzo Sani\textsuperscript{\affiliationA,\affiliationB}
\AND
Bill Marino\textsuperscript{\affiliationA}
\And
Preslav Aleksandrov\textsuperscript{\affiliationA}
\And
William F. Shen\textsuperscript{\affiliationA}
\And
Nicholas D. Lane\textsuperscript{\affiliationA,\affiliationB}
}

\begin{document}

{
\begingroup
\begin{figure}[t]
    \quad
    \begin{subfigure}{0.1275\textwidth}
        \includegraphics[width=\textwidth]{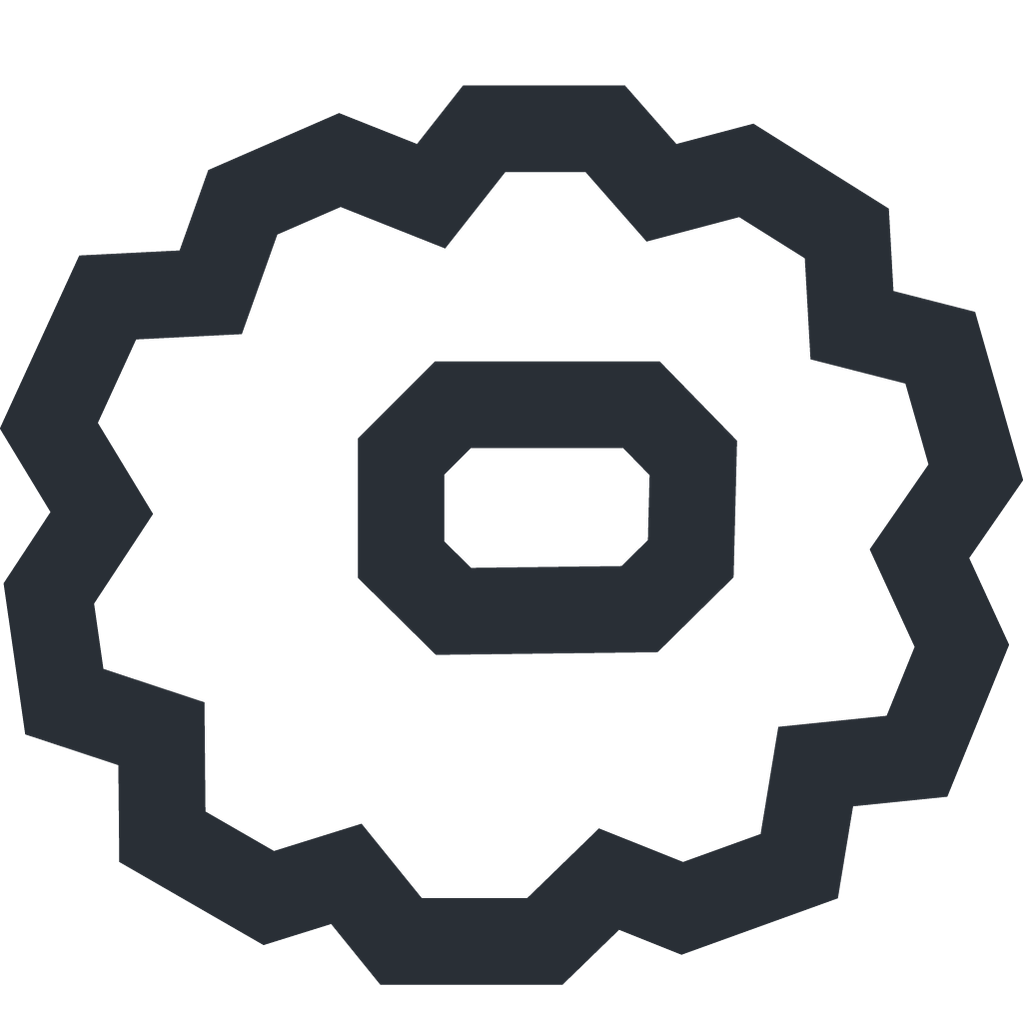}
    \end{subfigure}
    \hfill
    \begin{subfigure}{0.1\textwidth}
        \includegraphics[width=\textwidth]{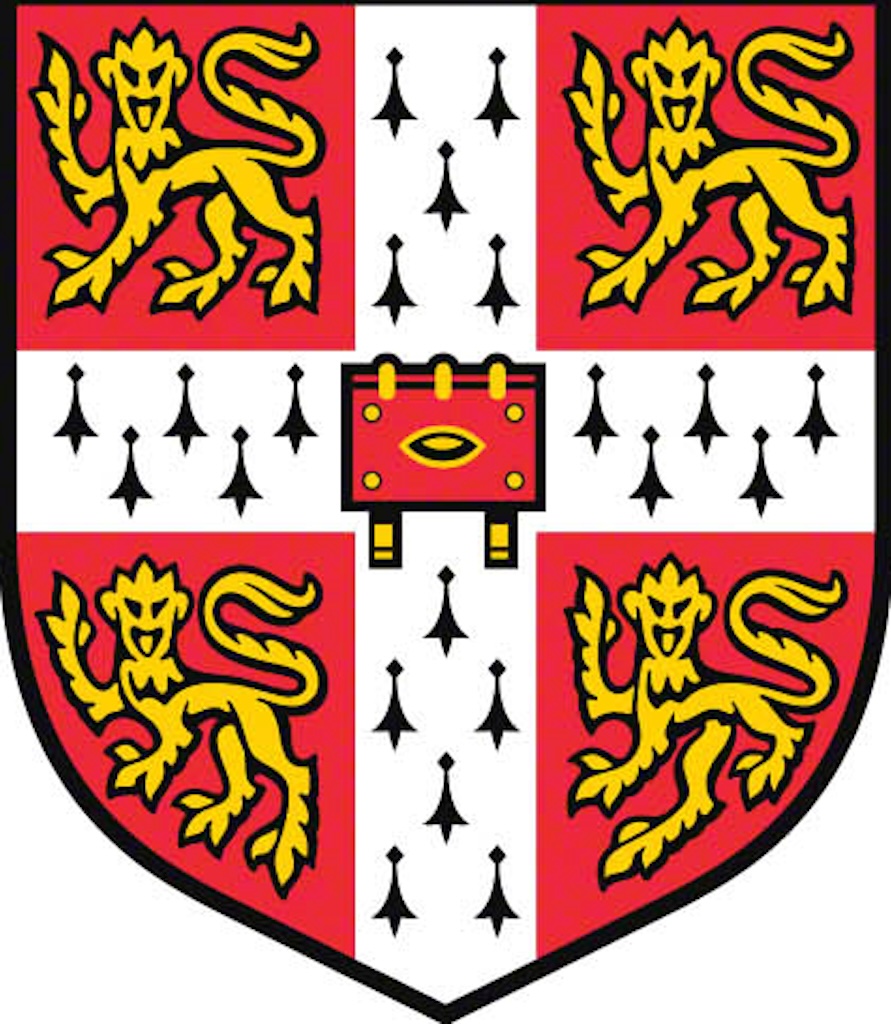}
    \end{subfigure}
    \hfill
    \begin{subfigure}{0.1275\textwidth}
        \includegraphics[width=\textwidth]{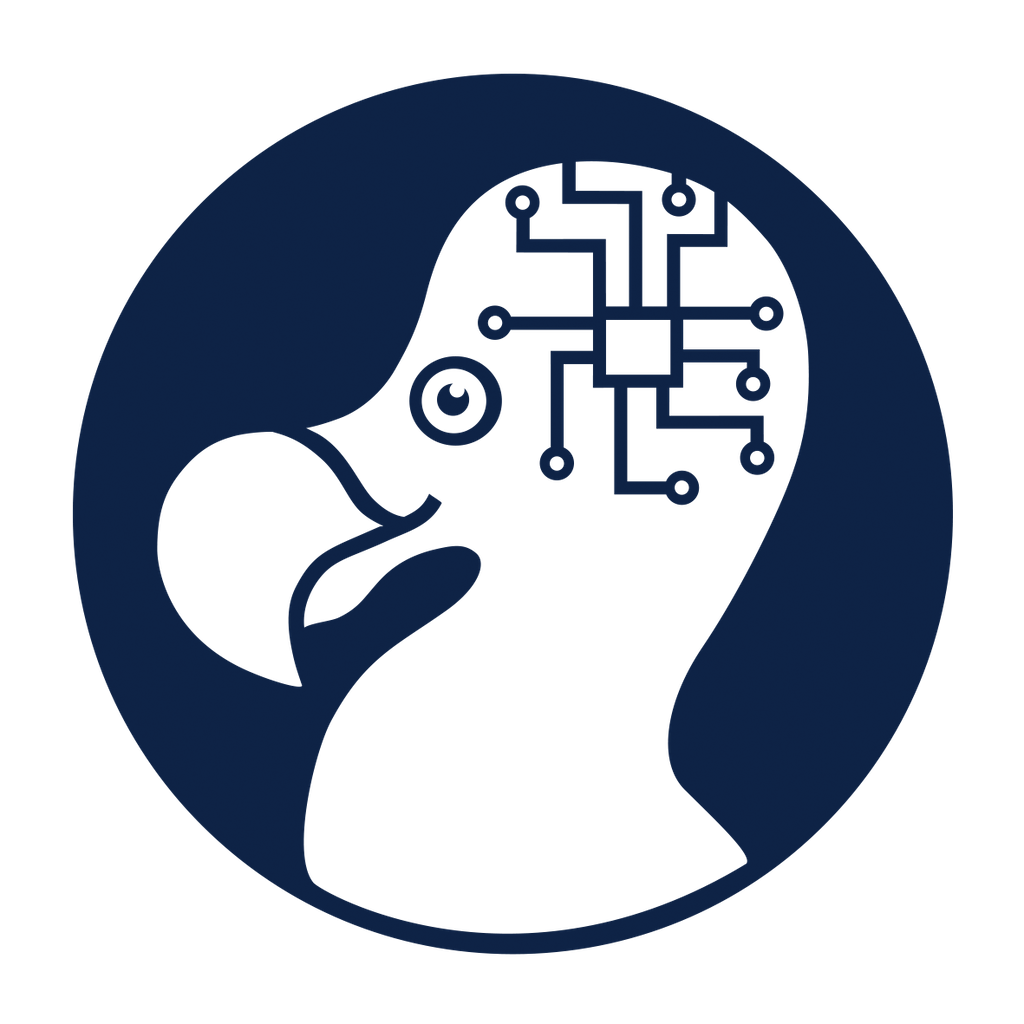}
    \end{subfigure}
\end{figure}
\endgroup
}
\setcounter{figure}{0}
\maketitle

\renewcommand{\thefootnote}{\myfnsymbol{footnote}}
\footnotetext[1]{Department of Computer Science and Technology, University of Cambridge}
\footnotetext[2]{Flower Labs}

\setcounter{footnote}{0}
\renewcommand{\thefootnote}{\fnsymbol{footnote}}

\begin{abstract}
  The reliance of language model training on massive amounts of computation and vast datasets scraped from potentially low-quality, copyrighted, or sensitive data has come into question practically, legally, and ethically. Federated learning provides a plausible alternative by enabling previously untapped data to be voluntarily gathered from collaborating organizations. 
However, when scaled globally, federated learning requires collaboration across heterogeneous legal, security, and privacy regimes while accounting for the inherent locality of language data; this further exacerbates the established challenge of federated statistical heterogeneity. We propose a Worldwide Federated Language Model Training~(\wrld) system based on \textbf{federations of federations}, where each federation has the autonomy to account for factors such as its industry, operating jurisdiction, or competitive environment. \wrld enables such autonomy in the presence of statistical heterogeneity via partial model localization by allowing sub-federations to attentively aggregate key layers from their constituents.
Furthermore, it can adaptively share information across federations via residual layer embeddings.
Evaluations of language modeling on naturally heterogeneous datasets show that \wrld outperforms standard federations by up to $1.91\times$, approaches the personalized performance of fully local models, and maintains these advantages under privacy-enhancing techniques.

\end{abstract}

\section{Introduction}
Contemporary Language Models~(LMs) are trained in data centers with numerous hardware accelerators~\citep{BLOOM,llama_3} on large scraped corpora~\citep{gpt3,DatasetsSizeSurvey}. This paradigm has restricted training to well-funded actors and brought concerns regarding centralization~\citep[sec. 4.7.1]{OpportunitiesAndRisksLLM}, copyright~\citep{NytVsOpenAi}, and the limited supply of data~\citep{WillWeRunOutOfData}. While small organizations may own valuable data, the per-batch synchronization constraints of stochastic gradient descent (SGD)~\citep{FSDP_ZeRO,FSDP_Pytorch} make collaboration impractical. Federated Learning~(FL) has been proposed as an alternative~\citep{DiLoCo, PositionPaper, LlmFLPlaceholder} due to its lower synchronization constraints, limited data movement, and ability to integrate privacy-enhancing techniques~\citep{DiffPrivacyFL,SecAggOG}. 

This work addresses the novel challenges that global-scale FL applications face. The first challenge is related to \textbf{federated governance} \citep{FLHealthGovernance}: how to create federated systems including participants with different legal, privacy, and security concerns or combine existing systems seamlessly without sacrificing performance. The second challenge is the \textbf{statistical heterogeneity} of naturally distributed data, as organizations may have datasets that differ in terms of language (e.g., for federations spanning multiple countries) and genres (e.g., 
publishers operating in different domains). Significant heterogeneity has been shown to slow model convergence or even cause divergence in standard single-federation settings~\citep{AdvancedAndOpenProblems,UnderstandingModelAveragingInFL,HeterogeneousFLSurvey}. This issue is worsened at a global level, as external factors~\citep{DatasetGeography} can implicitly create different clusters of data in the federation, making it necessary to optimize a participant's parameters with respect to both the global distribution and its peers. 

\begin{figure}[t]
    \centering
    \includegraphics[clip,width=0.8\columnwidth]{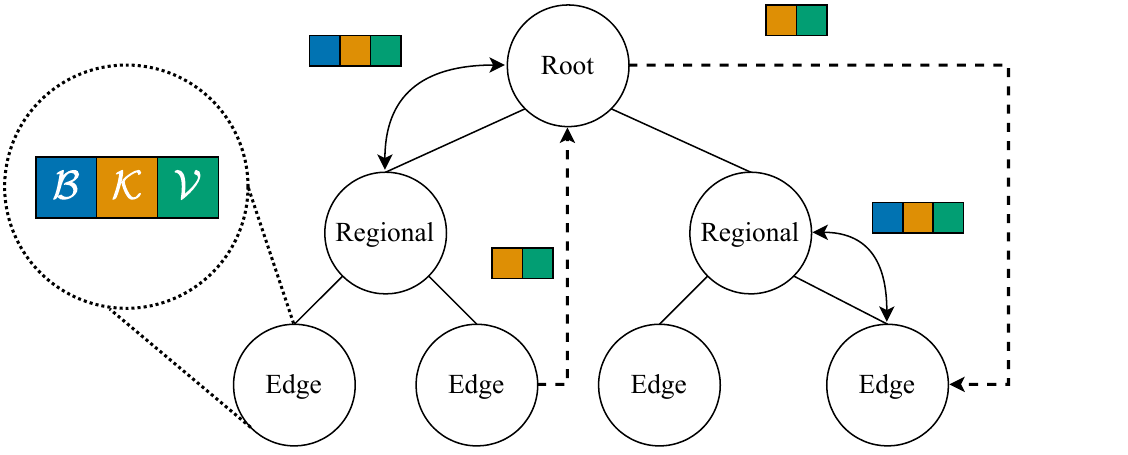}
    \caption[System Diagram]{\wrld federations exchange information in the form of models containing a backbone, personalized layers~($\mathcal{B,K,V}$), and lower-dimensional residual embeddings serving as keys and values~($\mathcal{K,V}$). While full models are exchanged between parents and children, residuals are dynamically routed to the most appropriate sub-federation to be used in attention-based aggregation.}\label{fig:TreeStructure}
\end{figure}

Our proposed Worldwide Federated Language Model~(\wrld) training system addresses both challenges by building \textbf{federations of federations}, as seen in \cref{fig:TreeStructure}. This brings several advantages:
\begin{enumerate}
    \item \textbf{Arbitrary Federation Structure:} \wrld allows each federation to adapt to its unique legal, privacy, and security requirements and also permits easy integration between previously isolated organizations, with differential privacy~\citep{DiffPrivacyFL} protocols. 
    \item \textbf{Partially-personalized Aggregation:} \wrld takes advantage of cluster relationships within federations by segmenting models into a \textit{backbone} serving as a shared feature extractor trained using standard FL algorithms~\citep{FedAvg} and partially-personalized \emph{key} layers relevant to a node. These \textit{key} layers are aggregated using attention over the keys of its peers. Results show that the aggregation procedure of \wrld allows it to achieve a perplexity lower by $11.5$-$45.3\%$ on a novel hierarchical partitioning of \textit{The Pile}~\citep{ThePile} dataset~(e.g, \cref{fig:DatasetStructureHFL3_4}) and on a language-partitioned \textit{mC4} dataset~\citep{mC4}. Furthermore, \wrld converges to similar performance when using differential privacy, whilst standard FL diverges completely.
    \item  \textbf{Cross-federation Information Sharing:} To address intra-federation heterogeneity, where a participant may contain data that is more similar to another sub-federation than to its peers, \wrld can route \emph{residual} parameters to the most relevant sub-federation in a communication-efficient fashion based on \textit{key} similarity. Including such residuals allows \wrld to partially maintain performance even when the underlying sub-federations contain dissimilar participants while requiring the transfer of a limited number of additional transformer blocks. We find 
    the last $1-3$ blocks to be sufficient.
   
\end{enumerate}

The novel optimization algorithm and residual information routing of \wrld tackle the challenges of global LM training without diverging far from the infrastructure of production systems~\citep{ScaleSystemDesign}. Thus, it provides an effective solution for training LMs across legal, socioeconomic, and cultural boundaries, diversifying the pool of available data away from its current geographic~\citep{DatasetGeography} concentration.

\section{Background}
Well-known performance scaling laws~\citep{OgScalingLaws,TrainingComputeOptimalLLMs} drive LM training. The resulting models~\citep{BLOOM,LLaMA,llama2,llama_3} have led to numerous decentralized downstream applications~\citep{hu2021lora, Quantized1bitLLMs,LSLlamaFineTunedLexicalSimplification,ParameterEfficientFinetuningLLAMAmedical,PEFTSurvey}.
However, since the Ring AllReduce algorithm~\citep{Horovod} used to synchronize the model state for every batch~\citep{FSDP_ZeRO,FSDP_Pytorch} requires high cross-GPU bandwidth, pre-training still requires large centralized datacenters~\citep{llama_3}.

\citet{DiLoCo} relax the communication requirements of training via Local SGD~\citep{LocalSGD}, a data-parallel technique synchronizing independent model replicas. Based on the observation that publically available data may not keep up with the demands of LM training~\citep{WillWeRunOutOfData}, future work~\citep{LlmFLPlaceholder} extended generative pre-training to Federated Learning (FL) settings and showed performance comparable to centralized training even for billion-scale models. This shift enables collaboration across geographically distributed actors with valuable data currently serving as passive data providers~\citep{OpenAISpringerDeal,OpenAINewsPublishersDeal}.   Given the potential of  FL to reach untapped data reservoirs, we are motivated by these results to tackle the challenges that emerge when applied at a \emph{global} scale.

\subsection{Global Federated Systems}
Standard federated training proceeds in rounds that involve broadcasting a shared set of parameters to all clients, local optimization of said parameters on private data, and then aggregation~\citep{FedAvg}. Its roots in the principle of data minimization~\citep{FedAvg,White_House_Report} make FL an attractive paradigm across international regulatory frameworks~\citep{WhiteHouseFL,PositionPaper}~(\cref{app:legal_context}). 
However, while it is communication-efficient and scalable, FL suffers from \textbf{statistical} and \textbf{system} heterogeneity, defined as clients having different data distributions, training speeds, or communication bandwidth. For example, \citet{LargeCohorts} and \citet{UnderstandingModelAveragingInFL} observe that training on more clients within a round brings diminishing returns due to near-orthogonal updates. These behaviors are analogous to the limitations of large-batch training ~\citep{LargeBatchGenGapSharpMinima}, indicating the need for more sophisticated aggregation strategies, unlike the Local SGD literature~\citep{DontUseLargeBatchesUseLocalSGD}, which assumes homogeneous data. Similarly, systems heterogeneity can result in varying client completion times~\citep{ScaleSystemDesign,PAPAYA}, causing stragglers and slow convergence.

Since high-quality language data is liable to run out within the decade~\citep{WillWeRunOutOfData}, the quality of data that FL makes available is also paramount. For example, obtaining data from regions with a reduced internet presence may alleviate the challenges associated with low-resource languages~\citep{LowResourceLanguagesSurvey,LowResourceNMTSurvey}. We argue that the impact of data and systems heterogeneity increases as the reach of the federated system grows, which causes governance to emerge as a new challenge.

Federated governance, hitherto considered primarily in the medical domain~\citep{BigDataCancer, FLmedicine}, requires determining how to create systems spanning actors with different legal, privacy, and security concerns and how to combine already existing federations. For example, unlike US-based organizations, training an LM on the private data of EU-based organizations requires abiding by the GPDR~\citep{GDPR}, which restricts cross-border data transfer, while mitigating bias ~\citep[Art. 10.2f,fa]{EuAiACt}~(see \cref{app:legal_context}). Although differential privacy (DP)~\citep{DiffPrivacyFL} or secure aggregation~\citep{SecAggOG} may enable collaboration across legally heterogeneous regions, not using them within a region when possible avoids computational and performance costs~\citep{SecAggOG,LightSecAgg,FastSecAgg,LearningDifferentiallyPrivateRNNs}. Similar hierarchical structures enable collaboration in platform ecosystems~\citep{CanWeAskYouToCollaborate,CreatingValueThroughCollaboratives}, OSS~\citep{OSSHierarchical}, and vertical FL on medical data\citep[sec. 2.4.1]{FlHealthcareDomainVertical}.

Privacy-enhancement techniques (PETs), such as DP \citep{DP_orig} or secure aggregation, have become popular in FL since client updates may leak private information~\citep{InformationLeakageFL}.
Researchers and practitioners consider DP the de facto method for alleviating privacy concerns in FL~\citep{DiffPrivacyFL,AdvancedAndOpenProblems}, as it provides statistical bounds on the leakage of private information.

\subsection{Related Work}
Previous works in standard federated learning have tackled aspects of data and systems heterogeneity relevant to our work~\citep{FLONNonIID, AdvancedAndOpenProblems,SalvagingFL,HeterogeneousFLSurvey}. However, they are unable to account for the complexities of building federations-of-federations.
Personalized Federated Learning~(PFL)~\citep{TowardsFairPrivacyPreservingFL} attempts to tackle the performance disparity between clients induced by the heterogeneity of client data by creating personalized models. The most common means of achieving this is a local adaptation (fine-tuning) of the federated model after training~\citep{SalvagingFL,ThreeApproachesMansour,FinetuningIsFineFL,FinetuningIsFineFL}. Split approaches such as those of \citet{FlWithPersonalisationLayers} and \citet{FedSplitBert} train models locally but only average and update the layers up to a cut layer $q$, with the intuition that earlier layers represent feature extractors. 

Another means of tackling data heterogeneity is clustering clients based on a similarity metric between the flattened model embeddings~\citep{AnEfficientFrameworkForClusteredFL,ClusteredFederatedLearningModelAgnostic,HierClustering, FedAMP}. However, such approaches reduce the compatibility of client models to a single scalar and do not consider the structure of the model. Clusters may also exist naturally based on characteristics like location or language, which become relevant if the clients and servers controlling them are geographically correlated. However, most previous works in this area have focused solely on communication efficiency rather than data heterogeneity~\citep{Client-Edge-CloudHierFL,Hier_Het_Cellular,HFELJointEdgeResource,ResourceEfficientHierAgg}. As such, they broadcast a global model rather than server personalization.

\citet{OptimalUserEdgeAssingmentHierFL} do consider scenarios where the data distribution of edge servers is taken into account and propose optimal user-edge assignment. While promising, their work assumes that edge servers have sufficiently low communication latency to efficiently train with FedSGD despite the original work of \citet{FedAvg} showing FedSGD to be up to two orders of magnitude slower than FedAvg in terms of convergence speed. Another work that addresses data heterogeneity within hierarchical FL is \citet{PersonalisedFederatedLearningForIntelligentIoT}; however, their framework focuses on well-known client-level fine-tuning strategies and never explicitly utilizes the hierarchical structure to enhance personalization.

\section{\wrld}

Given the interlocking nature of legal, privacy, and security concerns, we assume that organizations find it easier to collaborate with enterprises operating in the same geographic region, legal jurisdiction, or industry. Therefore, a key design objective for our system is facilitating these types of collaborations. In edge-device FL, the large number of potentially ephemeral participants with fine-grained preferences leads to a fully decentralized collaborative learning paradigm~\cite{AdvancedAndOpenProblems,FullDecentralizedFL} where the server-client federated network is replaced by a graph encoding client compatibility. For training LMs, where the participating entities are stable organizations, we argue that a \textbf{federation of federations} approach, as portrayed in \cref{fig:TreeStructure}, provides an attractive compromise between standard federated learning and decentralized learning. To optimize in this setting, \wrld takes advantage of a custom aggregation procedure and information-sharing mechanism.

\subsection{Partially-personalized Aggregation}

\begin{algorithm}[]
        \caption{$\mathrm{Fit}$: execution procedure for a given sub-federation. } \label{alg:WorldLM}
        \small
        \begin{onehalfspace}
        \begin{algorithmic}[1]
             \Require{Node id $q$, parent backbone $\mathcal{B}_p$, sequence of key layers $\mathcal{K}_p$}
             \Require{Downstream residuals for aggregation $\mathcal{D}_a$ for routing $\mathcal{D}_r^0$}
             \begin{green}
            
             \State{$\mathcal{B}^0, \mathcal{K}^0 \gets \mathrm{LoadModel}(q)$}
             \State{$\mathcal{U}^0 \gets \emptyset$} \Comment{Upstream residuals for parent}
             \If{$q \neq 0$} \Comment{If the current node is not root}
             \State{$\mathcal{Q,K,V} \gets \mathcal{K}^0, [\mathcal{K}^0, \mathcal{K}_p, \mathcal{D}_a], [\mathcal{K}^0,\mathcal{K}_p, \mathcal{D}_a]$}
             \State{$\mathcal{B}^0 \gets \mathcal{B}_p$} 
            
             \State{$\mathcal{K}^0 \gets \mathrm{Attn(\mathcal{Q,K,V})}$} 
             \Comment{Agg. node, parent and residual key layers} 
             \EndIf
             \end{green}
             \vspace{-0.4cm}
             \begin{white}
             \For{round $k \gets 0, \ldots, K-1$} 
             \EndFor
             \end{white}
             \vspace{-0.4cm}
            \begin{red}
            \Indent
            \vspace{-0.2cm}
            \State{$A^k, R^k \gets \mathrm{RouteResiduals}(\mathcal{D}_r^k, C)$} \Comment{Route using similarity to cached children $\mathcal{K}$}
            \State{$\mathrm{Train}(q, \mathcal{B}^{k}, \mathcal{K}^{k})$}  \Comment{Train node params sequentially or in-parallel}
            \For{child $c \in C_q$} \Comment{In parallel}
                \State{$\mathcal{B}_c^k, \mathcal{K}_c^k, \mathcal{U}_c^k \gets \mathrm{Fit}(c, \mathcal{B}^k, \mathcal{K}^k, A_c^k, R_c^k)$} \Comment{$\mathcal{U}_c^k$ are residuals sent upstream by the child}
                \State{$\Delta^{k}_{c} \gets \mathcal{B}_c^k - \mathcal{B}$} \Comment{Compute backbone pseudo-gradient}
            \EndFor
            \EndIndent
        \end{red}
            \begin{blue}
            \Indent
            \vspace{-0.2cm}
            \If{$C_q \neq \emptyset$}
            \State{$\Delta^k \gets \frac{1}{|C|}\sum_{c\in C}{\Delta^{k}_{c}}$} \Comment{Aggregate pseudo-gradients}
            \State{$\mathcal{B}^{k+1} \gets \mathrm{ServerOpt}(\mathcal{B}^k, -\Delta^k, k)$}   \Comment{Apply pseudo-gradient}
            \State{$\mathcal{Q,K,V} \gets  [\mathcal{K}^k_0,\ldots ,\mathcal{K}^k_{|C|}],\ldots,[\mathcal{K}^k_0,\ldots ,\mathcal{K}^k_{|C|}]$} \Comment{Pack key layers with the same index}
            \State{$\mathcal{K}^{k+1} \gets \mathrm{Attn(\mathcal{Q,K,V})}$}
            \State{$\mathcal{U}^{k+1}, \mathcal{D}_r^{k+1} \gets \mathrm{PartitionResiduals}(q, \mathcal{K}^{k+1}, V, \mathcal{U}^k, \mathcal{D}_r^{k})$} \Comment{Based on dissimilarity}
            \EndIf
            \EndIndent
            \end{blue}   
    \vspace{-0.4cm}
    \begin{white}
    \Return{$\mathcal{B}^{k}, \mathcal{K}^{k}, \mathcal{U}^{k}$}
    \end{white}
        \end{algorithmic}
        \end{onehalfspace}
\end{algorithm}

Federated learning can train powerful feature extractors beneficial to all \wrld participants due to its meta-learning properties\citep{REPTILE,PflModelAgnosticMetaLearning, lee2024fedl2p}. However, actors in the federated training process may hold heterogeneous data; for example, it may be in a different language or come from a separate domain~(e.g., one organization may publish news while another focuses on scientific publications). Thus, the feature extractor needs to be adapted to the data of each actor and sub-federation, which requires a departure from the standard FL objective. Thus, \wrld takes inspiration from split-learning techniques~\citep{FlWithPersonalisationLayers,FedSplitBert} and partitions the model into a backbone $\mathcal{B}$, comprising the majority of the model's parameters, and a set of partially personalized key layers $\mathcal{K}$ specific to each node. In practice, we expect the key layers to include the final $1$-$3$ transformer blocks.

The backbone parameters $\ba$ are trained using FL aggregation algorithms such as FedAvg~\citep{FedAvg} or FedOPT~\citep{FedOPT}. As shown in \cref{alg:WorldLM}~(\colorbox{redc}{L.$8-12$}), for each round $k$, the server broadcasts $\ba$ to its children $c \in C_q$. When a child begins execution, it replaces its previous backbone with the one received from the parent~(\colorbox{greenc}{L.$5$}), and then executes its own sub-federation~(\colorbox{redc}{L.$10-12$}) and/or local training~(\colorbox{redc}{L.$9$}) and computes pseudo-gradients $\Delta_{c}^k$. These pseudo-gradients are then aggregated and applied to the backbone of the server via $\mathrm{ServerOpt}$~(\colorbox{bluec}{L.$15$}). It is important to note that local training~(\colorbox{redc}{L.$9$}) can be executed either \textbf{in-parallel} with the node model treated as a client or \textbf{sequentially}. Similarly, when observing the performance of a hierarchical FL approach we distinguish between sequential and parallel steps as parallel steps are averaged across clients. 
For this work, we choose the \textbf{sequential} approach, meaning that for the first level of the hierarchy, the root executes by itself, whilst for the final level, all leaves are executing \textbf{simultaneously}.

Unlike traditional personalization approaches, \wrld optimizes the key layers $\ke$ while considering the models of its parent and children. It achieves this by aggregating each layer within $\ke$ using an attention mechanism~\citep{AttentionIsAllYouNeed,AttentiveAggregation} where the parameters of that given layer across the nodes of the federation represent the queries, keys, and values. To exploit the locality of data and the similarity of nodes within the same sub-federation, the attention mechanism is applied within a local context, either with respect to the parent of a given node or its children.  This procedure allows \wrld to handle statistical heterogeneity when the nodes within a sub-federation have a similar data distribution.

\subsection{Cross-Federation Information Sharing}

The attention-based aggregation of a node's children~(\colorbox{bluec}{L.$17$}) will reduce to simple unweighted averaging when a cluster relationship does not exist in a sub-federation. In contrast, the aggregation between a node and its parent~(\colorbox{greenc}{L.$6$}) will focus almost exclusively on the keys of the node and thus approach a personalization-layer strategy. Similarly, if a given node has a data distribution that differs significantly from its peers, its key layers will be highly dissimilar and ignored in the aggregation.

However, being dissimilar to its peers does not preclude a node from containing valuable data for another sub-federation. Inspired by residual connections~\citep{ResNet}, we introduce the concept of residual key layers, which are dynamically chosen based on their dissimilarity to peers and routed to a more appropriate sub-federation. For each layer of its post-aggregation key $\ke$, a node selects at most $\nu_\mathcal{K}$ of its children's layers with the lowest similarity score~(\colorbox{bluec}{L.18}), e.g., by the dot product. It then sends these residual layer embeddings to the highest permissible level of the federation as dictated by the participants' legal, security, and privacy concerns. This server then routes the residuals to its highest-similarity children based on a cached set of keys from the previous round~(\colorbox{redc}{L.8}) by placing them in $R^k$. When the residuals reach an edge node, they are placed in $A^k$ and later aggregated. Aggregating in the edge nodes~(\colorbox{greenc}{L.$4-6$}) has the benefit of diffusing information into the entire sub-federation right after local training.

\section{Experimental Design}

\begin{figure}[t]
    \centering
    \includegraphics[clip,width=0.8\columnwidth]{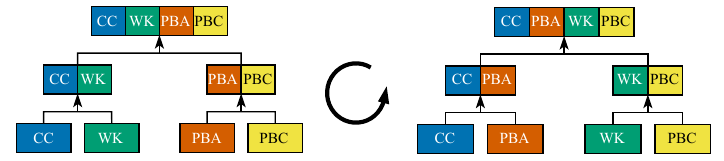}
    \caption[System Diagram]{Data-perspective upon a hierarchical dataset constructed from The Pile~\citep{ThePile}. The LHS contains two naturally heterogeneous and quantity-skewed groupings of data sources, corresponding to organizations accessing data from the internet or the medical domain. We construct such groupings using the internet-based Common Craw~(\textbf{CC}) and Wikipedia~(\textbf{WK}) versus the medial data of PubMed Abstracts~(\textbf{PBA}) and PubMed Central~(\textbf{PBC}).  To test the effectiveness of \wrld when such a cluster relationship is absent, we swap the position of the two smaller datasets.} \label{fig:DatasetStructureHFL3_4}
\end{figure}

Given the recent emergence of federated generative pre-training and the lack of benchmark datasets, we evaluate \wrld on tasks that approximate realistic scenarios for its application: (a) organizations in different industries collaborating to train an LM despite holding different genres of text, (b) organizations trying to train an LM despite holding data in different languages. 

To simulate organizations holding different genres of text, we partition \emph{The Pile}~\citep{ThePile} into its constituent heterogeneous datasets and construct federations of federations by bottom-up building different mixtures of datasets. For example, as seen in \cref{fig:DatasetStructureHFL3_4}, if the children of a node hold data from the \textbf{Pile Common Crawl}, \textbf{Wikipedia}, \textbf{PubMed Central}, and \textbf{PubMed Abstracts} datasets, then the node itself will hold data from them, proportionate to their size. For this dataset, we use the common \emph{gpt-neox-20b} English tokenizer also adopted by \citet{LlmFLPlaceholder}

To simulate geographically distributed systems, we use a subset of Multilingual Colossal Common Crawl~(\emph{mC4})\citep{mC4}, covering high and low-resource languages~\citep{LowResourceLanguagesSurvey}. We then build sub-federations based on language families similarly to \emph{The Pile}. Given the larger vocabulary size~\citep{mC4} and consequent model size, we use a single federated structure partitioned with the high-resource French~(\textbf{FR}) and Italian~(\textbf{IT}) on one side and the lower-resource Bulgarian~(\textbf{BG}) and Ukrainian~(\textbf{UK}) on the other. For IID experiments, we use the standard Cleaned Colossal Common Crawl~(\emph{C4}) English dataset, partitioned into equal-sized shards with the same tokenizer as \emph{The Pile}.

\subsection{Tasks}

\begin{table}[]
\centering
\caption{Hyperparameters for \wrld, the federated learning rate $\boldsymbol{\eta_s}$ and momentum $\boldsymbol{\mu_s}$~\citep{FedMOM} are applied by a standard FL server or \textbf{each} server of \wrld, $\boldsymbol{|\mathcal{K}|}$ is the number of blocks used for the key in \wrld. In contrast, $\boldsymbol{\nu_\mathcal{K}}$ is the number of layers we select for each residual across all clients. Finally, $\mathbf{S_C}$ are the parameters of the learning rate scheduler synchronized across \textbf{sequential} steps.}
\label{tab:fl_hyperaparams}
\begin{tabular}{@{}rcccccc@{}}
\toprule
    \textbf{Model (size)} &
    \textbf{\#Rounds} &
    $\boldsymbol{\eta_s}$ &
    $\boldsymbol{\mu_s}$ &
    $\boldsymbol{|\mathcal{K}|}$ &
    $\boldsymbol{\nu_\mathcal{K}}$ &
    $\mathbf{S_C(\boldsymbol{\alpha},~\boldsymbol{\eta_{max}},~T)}$ \\ \midrule
\textbf{English~($\mathbf{75}$M)} & $12$ & $0.2$ & $0.9$ & 1 & 1 & $(10^{-2},~8\times10^{-4},~3\times10^3)$ \\
\textbf{English~($\mathbf{125}$M)} & $21$ & $0.2$ & $0.9$ & 3 & 1 & $(10^{-2},~6\times10^{-4},~5\times10^3)$ \\
\textbf{Multi~($\mathbf{250}$M)} & $12$ & $0.2$ & $0.9$ & 1 & 1 & $(10^{-2},~8\times10^{-4},~5\times10^3)$ \\ \bottomrule
\end{tabular}
\end{table}

For all experiments, we use decoder-only transformers~\citep{gpt3} for the language modeling task. Given that we are concerned with partially personalized FL, we compare the personalized \textbf{local} perplexity of \wrld against three relevant baselines representing alternative scenarios. First, we compare against our closest alternative, standard FL with momentum~\citep{FedMOM,DiLoCo,LlmFLPlaceholder}, which has no hierarchical structure and would be difficult to integrate across legally and organizationally heterogeneous participants given the aforementioned governance issues. Second, we compare against independently trained fully personalized local models which do not share information, thus being heavily constrained by the local dataset size. This may result in heavy memorization due to overfitting to the limited amount of data of one organization. Third, we compare against centralized training of a global model on all the node's data pooled together, irrespective of privacy concerns. The centralized baseline approximates the upper bound performance for a global model.

\textbf{Language Modelling:} We focus on an underrepresented segment: groups having access to a meaningful but not sufficient number of GPUs. For example, organizations holding $8$ NVIDIA A40s~(as we use in this work). In such scenarios, while billion-scale LMs may be out of reach due to sheer VRAM and bandwidth requirements, medium models may be practically useful~(e.g., by providing highly relevant embeddings) without possessing the externalities of pre-trained models. Thus, given the hardware, we use three scales of models~($75$M, $125$M, and $250$M). Our federated training uses the parameters shown in \cref{tab:fl_hyperaparams} while the local parameters are available in \cref{tab:lm_hyperaparams}. We compare models that have executed for a given number of \textbf{sequential} steps, the step-wise execution we chose for \wrld~(\cref{alg:WorldLM}) implies that each level of a hierarchy~(\cref{fig:TreeStructure}) trains in parallel; however, the levels execute sequentially relative to each other in \textbf{three} stages.

\textbf{Privacy and Security:} To validate the effectiveness of \wrld for enhancing privacy and security, we simulate differentially private training, in which the leaves of a hierarchy contain potentially sensitive information and thus use \emph{DPFedAvg}~\citep{DiffPrivacyFL,DPClipping} instead of standard averaging. This requires gradient clipping and the injection of Gaussian noise. To this end, we assumed that the \textbf{Pile-CC} and \textbf{WK} leaf clients of \textit{The Pile} require DP and inject noise with $\sigma=0.5$ while clipping their gradients to the median $l_2$ norm of the previous round~(for the first round we use a bound of $1.0$ as done for centralized ML~\citep{BLOOM}). Given the hierarchical training of \wrld and the personalization involved, we postulate that it can more effectively account for the DP noise.

\FloatBarrier
\section{Evaluation}\label{sec:evaluation}

\begin{table}[]
\centering
\caption{Language modeling personalized performance~(over \textbf{local} client test sets) of \wrld in terms of perplexity (the lower, the better). We compare against standard FL, fully local, and centralized models. The latter is trained on the union of all local training sets. \wrld outperforms standard FL across all Non-IID dataset partitions, reaching a perplexity almost twice lower for \emph{mC4}. While fully local models may seem to perform well, they are unlikely to generalize out of distribution and are prone towards divergence due to small dataset sizes, as seen in \cref{fig:fed:pre_train_mc4_ppl}.
}
\label{tab:pre-train-results}
\begin{tabular}{@{}rrcccc@{}}
\toprule
 & & \multicolumn{2}{c}{Collaborative} & \multicolumn{2}{c}{Non-Collaborative} \\\cmidrule(rl){3-6}
\textbf{Dataset}&\textbf{Model} & \textbf{\wrld} &  \textbf{FL} &\textbf{Local}     & \textbf{Centralized}\\
\cmidrule(r){1-2}\cmidrule(rl){3-4}\cmidrule(rl){5-6}
\textbf{Pile}     & \textbf{75M}   & $\phantom{1} \mathbf{73.82 \pm 44.18}$ & $107.31 \pm 52.50$ & $ \mathbf{40.66 \pm 25.28 }$ & $85.81 \pm 24.42 $   \\
\textbf{Pile}     & \textbf{125M}  & $\phantom{1}  \mathbf{48.34 \pm 32.41}$ & $\phantom{1}53.92 \pm 24.24$  & $ \mathbf{24.83 \pm 12.47} $ & $29.61 \pm 13.17$    \\
\textbf{MC4}      & \textbf{250M}  & $\phantom{1}  \mathbf{80.47 \pm 68.53}$ & $153.27 \pm 95.47$ & $ \mathbf{45.47 \pm 31.13}$  & $72.21\pm 49.78$     \\
\textbf{C4} & \textbf{75M} & $167.31 \pm \phantom{1}2.92$ & $\mathbf{145.32 \pm \phantom{1}3.53}$ & N/A & $67.01 \pm \phantom{1}1.67$ \\ \bottomrule
\end{tabular}
\end{table}

\begin{figure}[]
    \centering
    \subfloat[]{\includegraphics[width=0.5\textwidth]{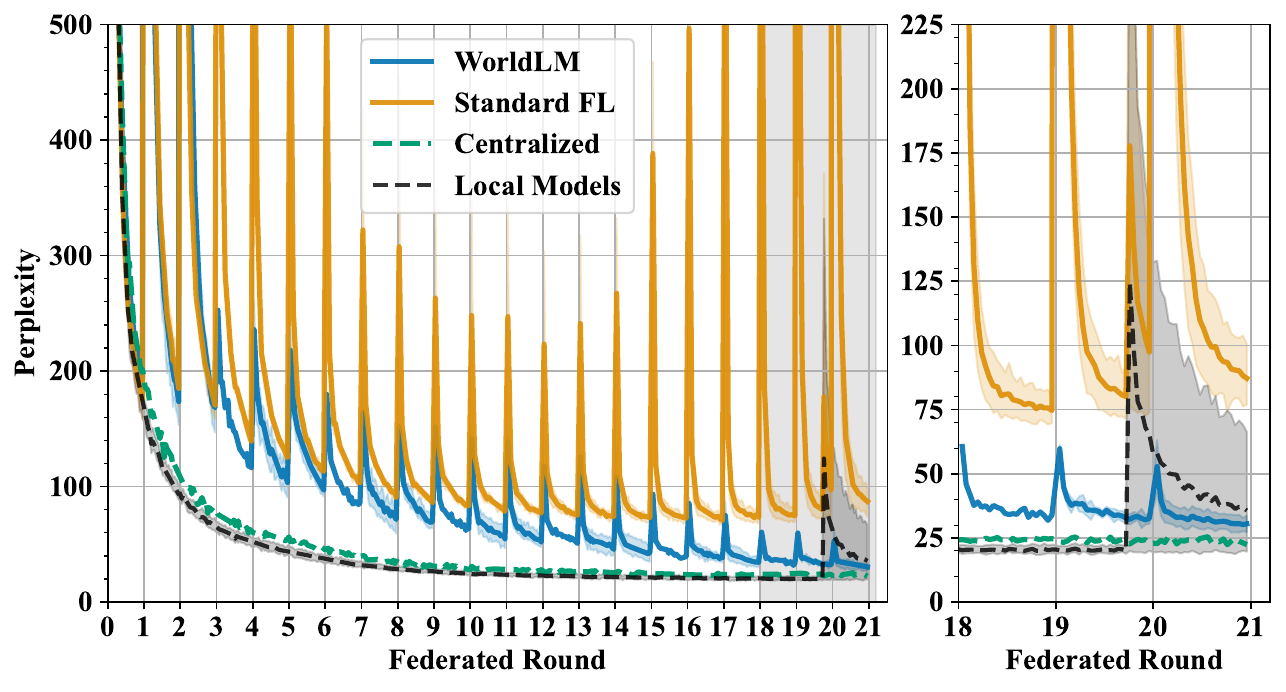}} 
    \subfloat[]{\includegraphics[width=0.5\textwidth]{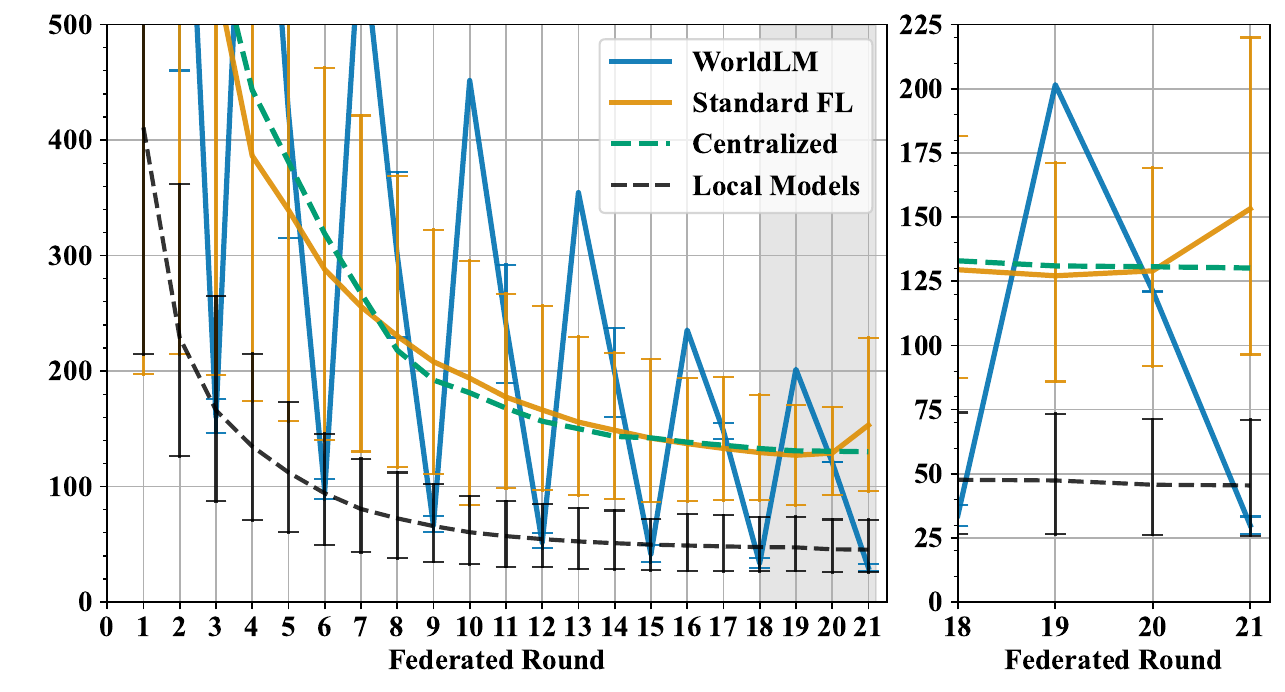}}\\
    \caption{\wrld training~(a) and validation~(b) \textbf{local} performance of the $250$M multilingual model trained on a three-level heterogeneous partitioning of \emph{mC4} constructed analogously to \cref{fig:DatasetStructureHFL3_4} and composed of the high-resource \emph{Italian} and \emph{French} languages on one side with the lower-resource \emph{Ukrainian} and \emph{Bulgarian} on the other. While standard FL stops improving after round $15$, \wrld reaches a performance close to the local and centralized models. The spike in local model perplexity at the end is due to eventual overtraining on the small \emph{Bulgarian} and \emph{Ukrainian} datasets.}
    \label{fig:fed:pre_train_mc4_ppl}
\end{figure}

Our results for the language modeling task~(\cref{tab:pre-train-results}) show that \wrld offers the desired compromise between global performance and local personalization. \wrld is capable of outperforming standard FL in terms of \textbf{local} personalized performance for datasets that exhibit statistical heterogeneity, while obtaining similar performance on the IID \textit{C4} dataset. Furthermore, it approaches a fully-centralized baseline for the $75$M model trained on \textit{The Pile} and the $250$M model trained on \textit{mC4}. As can be seen in \cref{fig:fed:pre_train_mc4_ppl}, the \textbf{partially-personalized aggregation} of \wrld allows it to approach the performance of a centralized model despite data never leaving each organization's premises. This holds the promise of international collaboration towards the federated training of LLMs. Since \wrld executes in three stages~(\cref{alg:WorldLM}) for a $3$-level hierarchy, \cref{fig:fed:pre_train_mc4_ppl} exhibits temporary \textbf{perplexity spikes} in the convergence curve~(e.g, the triangle formed by rounds $18-21$ in \cref{fig:fed:pre_train_mc4_ppl}). We chose to show all three stages since they correspond to different \textbf{sequential} steps. Thus, the root executes by itself while all nodes on a level execute together. For comparisons not dependent on the stage, consult \cref{tab:pre-train-results}.

\begin{figure}[]
    \centering
    \subfloat[]{\includegraphics[width=0.5\textwidth]{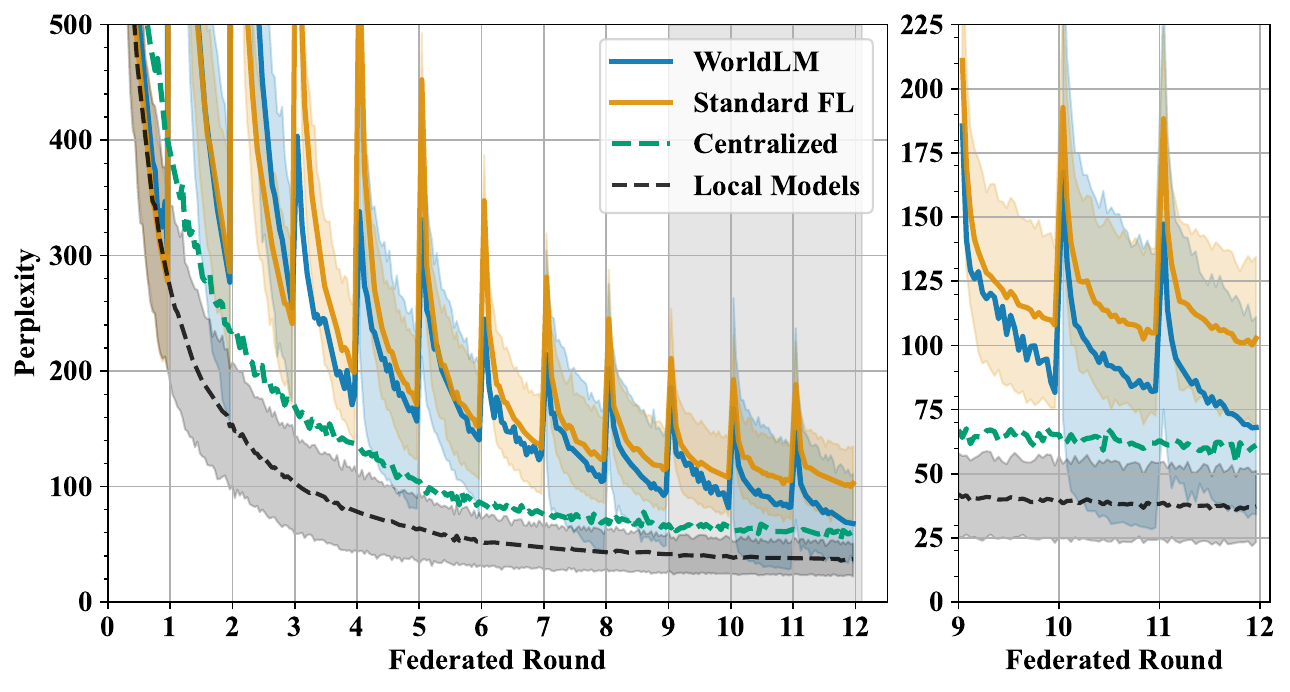}} 
    \subfloat[]{\includegraphics[width=0.5\textwidth]{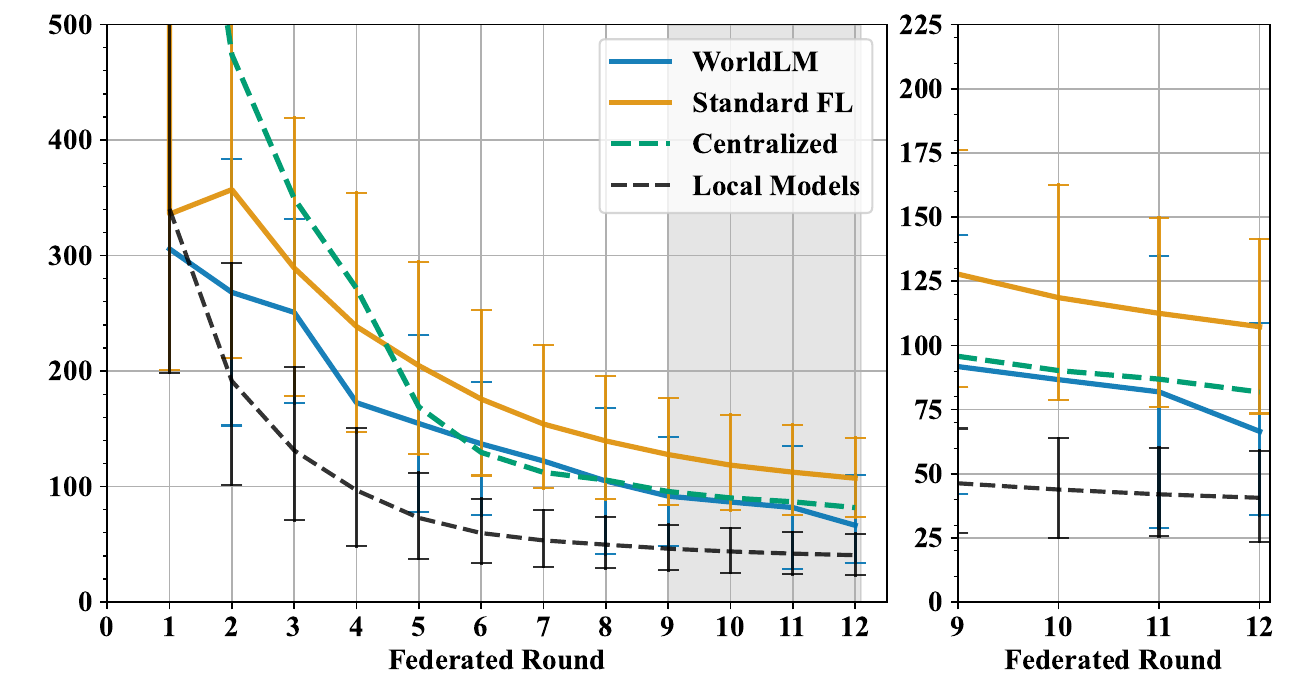}}\\
    \subfloat[]{\includegraphics[width=0.5\textwidth]{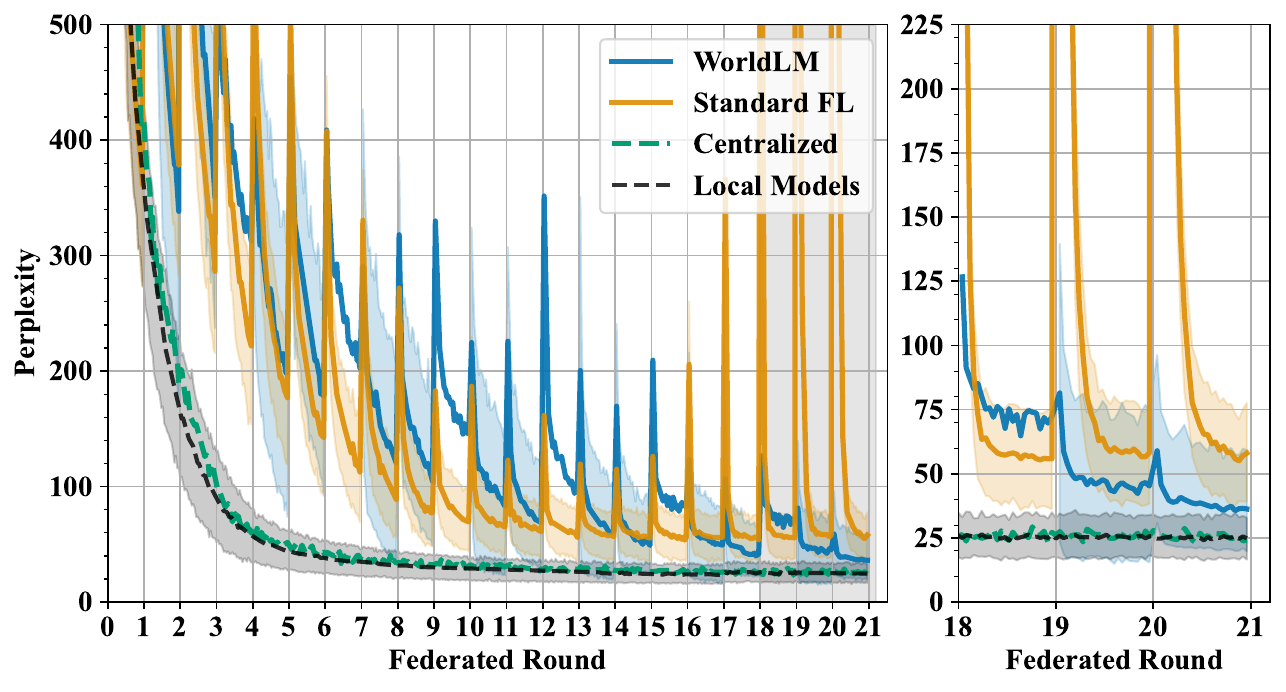}} 
    \subfloat[]{\includegraphics[width=0.5\textwidth]{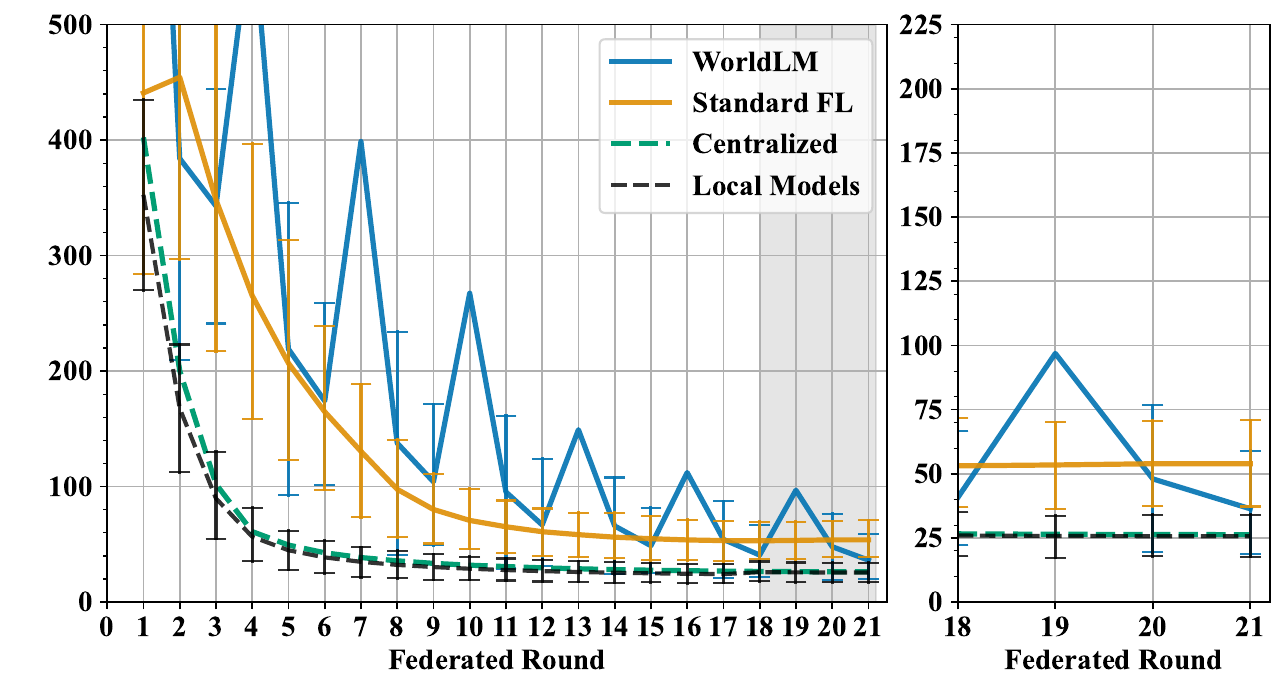}}\\
    \caption{\wrld training and validation performance of the $75$M~(a, b) and $125$M(c, d) English models on a three-level heterogeneous partitioning of \emph{The Pile}~(\cref{fig:DatasetStructureHFL3_4}). While the hierarchical approach makes steady progress due to its attention-based aggregation and partial personalization, standard FL struggles to converge due to data heterogeneity. Crucially, the performance of \wrld approaches that of the centralized model and partially overlaps with overfitted local models. }
    \label{fig:fed:pre_train_the_pile_ppl}
\end{figure}

Similar results can be observed on our heterogeneous partition of \textit{The Pile} in \cref{fig:fed:pre_train_the_pile_ppl}. However, the smaller heterogeneity between English texts causes a smaller gap between \wrld and standard FL. We use the $75$M model and \textit{The Pile} to evaluate the resilience of \wrld to differential privacy and changes in the data hierarchy~(\cref{fig:DatasetStructureHFL3_4}) due to being compute-limited. 

\begin{figure}[]
    \centering
    \subfloat[]{\includegraphics[width=0.5\textwidth]{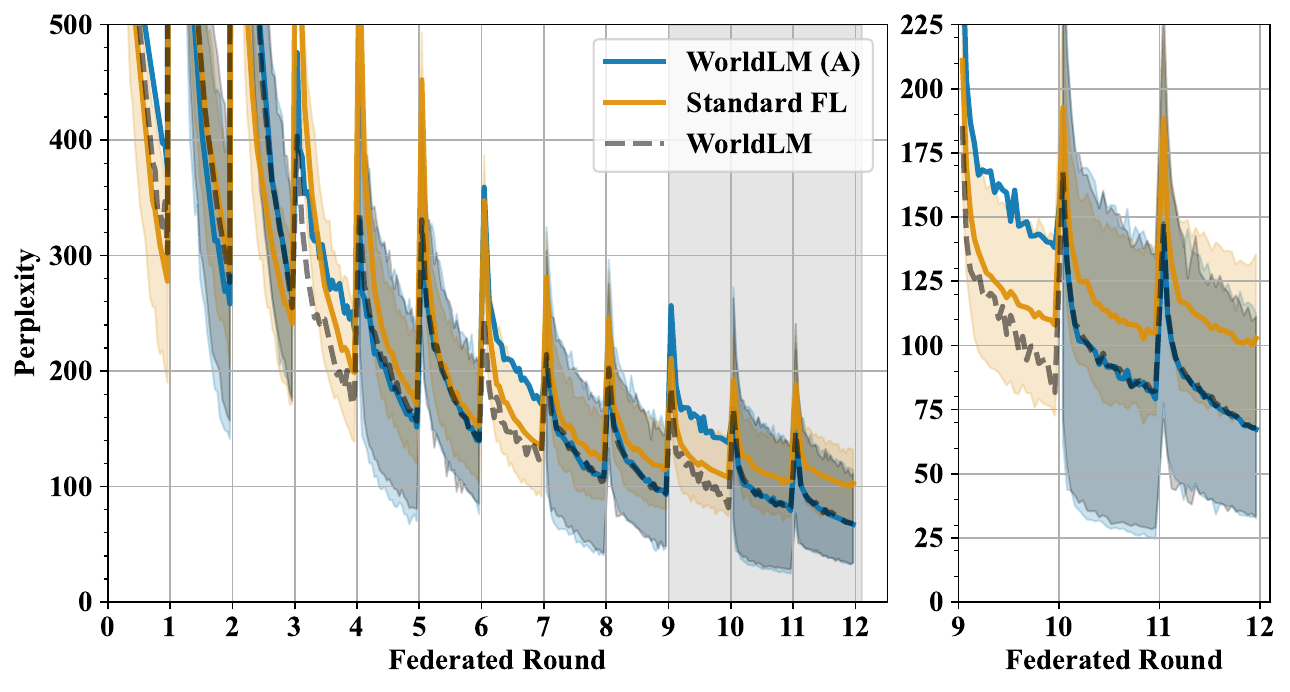}} 
    \subfloat[]{\includegraphics[width=0.5\textwidth]{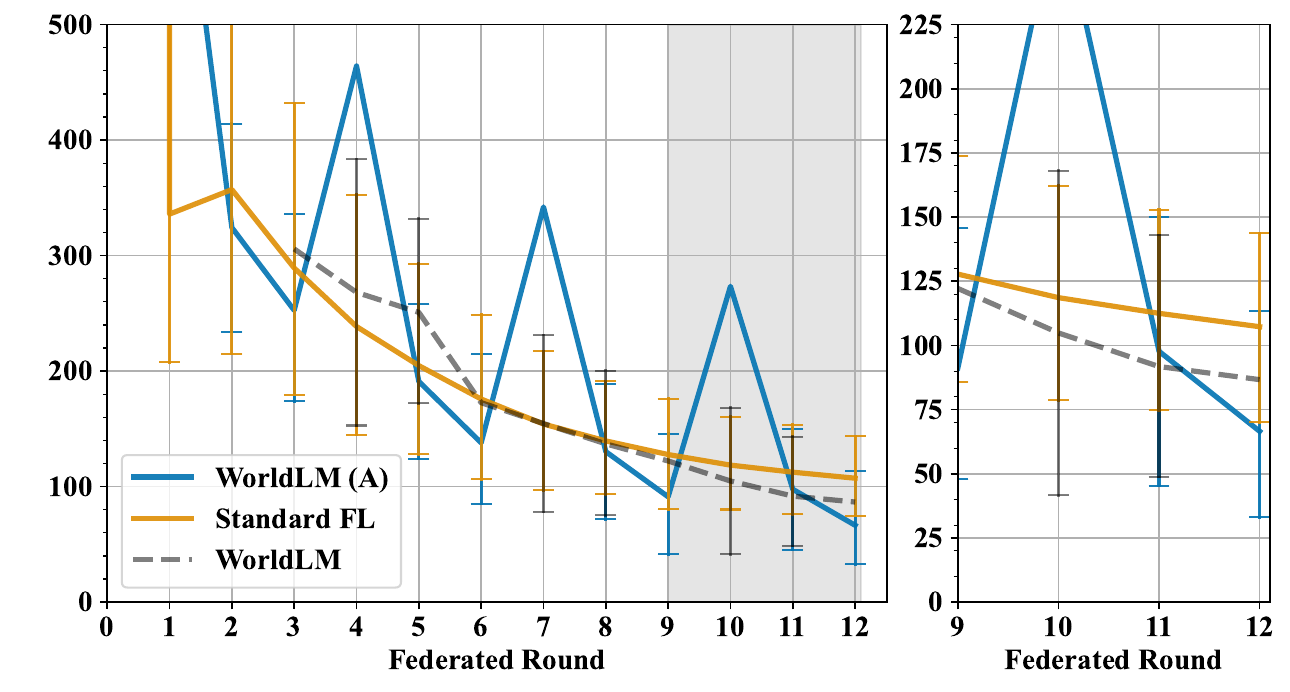}}\\
    \caption{The impact of swapping \textbf{WK} with \textbf{PBA} for \textit{The Pile} dataset~(see \cref{fig:DatasetStructureHFL3_4}) four our $75$M model. The swap results in the root node having worse performance~(rounds $6-7$ and $9-10$) due to being unable to reconcile the conflicting update directions from its sub-federations. Despite this fact, the personalized $\ke$ layers of the other nodes adjust the backbone $\ba$ to their local distribution. The \textbf{cross-federation information sharing} also permits the parameters of similar nodes~(e.g, \textbf{PBA} and \textbf{PBC}) to jointly optimize their keys, preserving overall \textbf{local} test-set performance.}
    \label{fig:fed:alt}
\end{figure}

We evaluate the robustness of \wrld to differential privacy and using an \textbf{alternative} arrangement of \textit{The Pile} which does not contain an inherent cluster relationship. The main limitation of \wrld is an imperfect ability to reconcile hierarchies with a high degree of intra-federation heterogeneity. As can be seen \cref{tab:dp-results} and \cref{fig:fed:alt} swapping the data of \textbf{WK} with \textbf{PBA}~(\cref{fig:DatasetStructureHFL3_4}) results in worse performance compared to standard FL due to the $\ke$ layers of the root being unable to agree during attention-based aggregation. Crucially, as can be observed from \cref{fig:fed:alt}, the personalized layers of the other nodes, together with the residual mechanism of \wrld allow them to maintain performance despite this decrease for the root. Consequently, for a majority of the  participating organizations \wrld serves as a superior alternative to FL from a personalization perspective.

\begin{figure}[]
    \centering
    \subfloat[]{\includegraphics[width=0.5\textwidth]{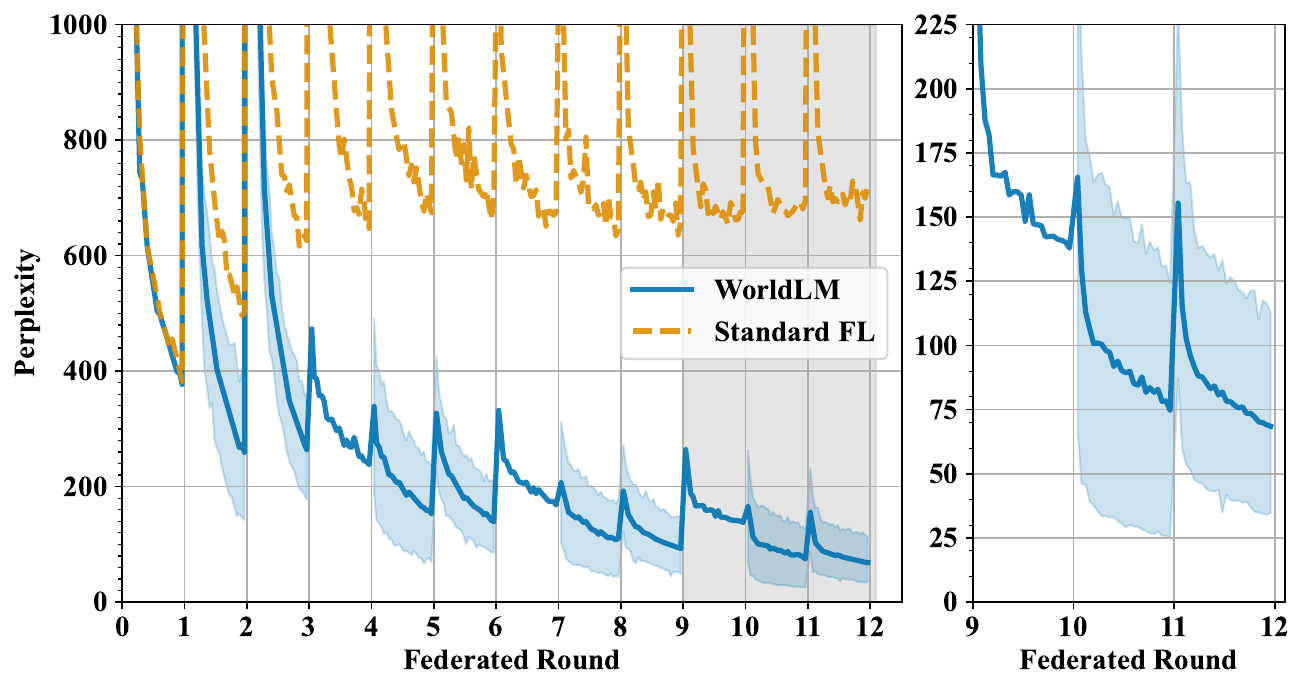}}
    \subfloat[]{\includegraphics[width=0.5\textwidth]{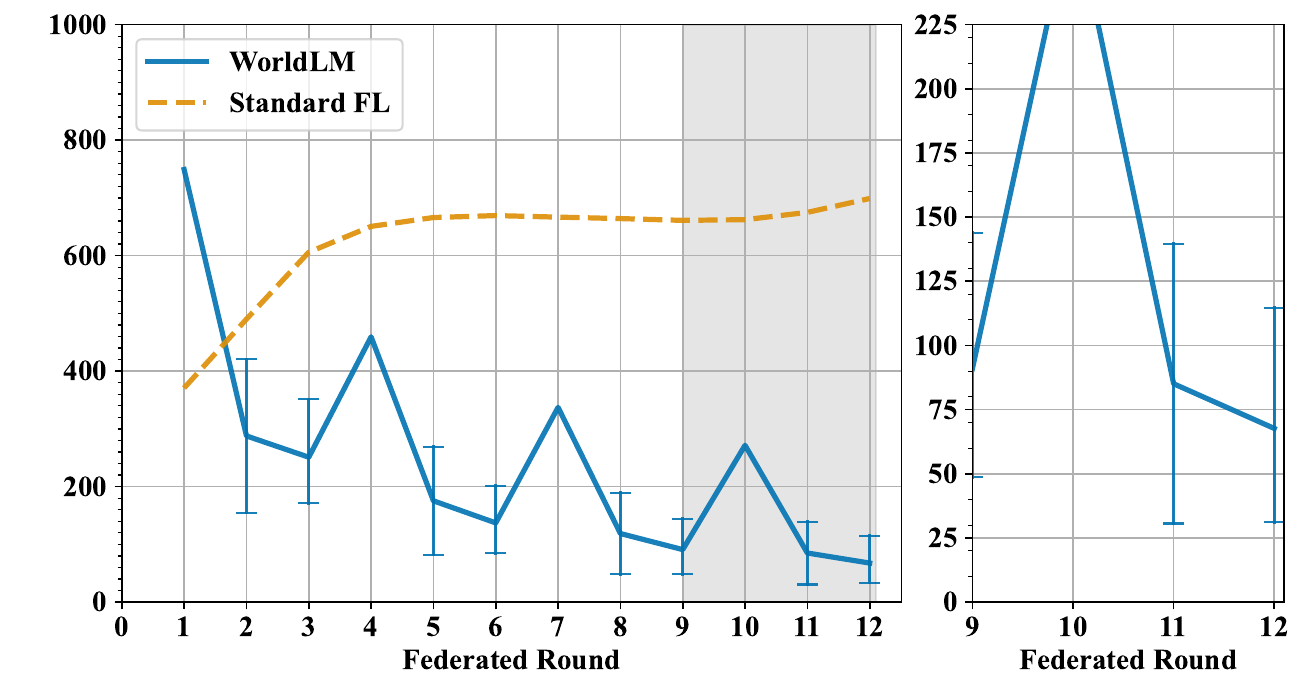}}\\
    \caption{The impact of differential privacy on \wrld performance, with the $75$M model on \emph{The Pile}, versus standard FL when using Gaussian noise with $\sigma=0.5$ and clipping client pseudo-gradients to the median pre-clip norm of the previous round. \wrld is able to stabilize the training and validation performance~(a,b) when applying DP over the leaf \textbf{CC} and \textbf{WK} clients, unlike FL which diverges.  While the attentional aggregation and parent-regularization maintain \wrld performance, standard FL allows the noise of even two clients to diverge the entirety of the aggregation procedure. }
    \label{fig:fed:dp}
\end{figure}

\cref{tab:dp-results} shows that \wrld is generally more robust than standard FL to the gradient clipping and noise that DP injects into the models of two leaf nodes. Standard FL diverges immediately due to its inability to suppress the impact of DP on the global model.  The personalized keys of \wrld, on the other hand, can ignore the impact of the noise entirely, as seen in \cref{fig:fed:dp}. Furthermore, the additional per-level momentum mechanism of \wrld allows it to stabilize the backbone training. This indicates that additional means of accounting for noise may be highly beneficial for standard FL approaches as well. For example, first pre-training on non-DP clients.

\begin{table}[]
\centering
\caption{Results evaluating privacy and robustness. \wrld is highly resilient to DP with $\sigma=0.5$ being applied over two of its leaf participants due to its aggregation procedure and because they start from a better initialization due to parent training. By contrast, the high amount of noise causes the entire global model of standard FL to diverge, without any personalized portions to maintain the performance of the non-DP members. Regarding robustness, \wrld suffers a decrease in performance once the cluster relationship in ~\cref{fig:DatasetStructureHFL3_4} is broken in \textbf{Pile~(A)} but remains competitive.}
\label{tab:dp-results}
\begin{tabular}{@{}rcccc@{}}
\toprule
\textbf{Method} &
  \textbf{Pile} &
  $\mathbf{\mathrm{DP}_{\mathrm{CC},\mathrm{WK}}}$ &
  \textbf{$\mathbf{\mathrm{DP}_{\mathrm{PBC},\mathrm{PBA}}}$} &
  \textbf{Pile~(A)} \\ \midrule
\wrld &
  $\phantom{1} \mathbf{73.82 \pm 44.18}$ &
  $ \mathbf{101.78 \pm \phantom{1}88.48}$ &
  $\phantom{1}\mathbf{103.68 \pm \phantom{1}90.53} $ &
  $140.05 \pm 100.52$ \\
\textbf{FL} &
  $107.31 \pm 52.50$ &
  $724.56 \pm251.89$ &
  $724.24 \pm250.98$ &
  $\mathbf{107.31 \pm \phantom{1}52.50}$ \\ \bottomrule
\end{tabular}
\end{table}

\FloatBarrier
\section{Conclusion}
\wrld supports the extension of federated learning~(FL) to the challenging setting of worldwide optimization of language models~(LMs). Our results indicate that systems based on \textbf{federations-of-federations} can compete with standard FL and centralized optimization for the medium-sized LMs affordable to small organizations and groups, given their hardware. Our results show that \wrld can outperform standard FL under realistic federated topologies and data distributions constructed using naturally heterogeneous datasets. Furthermore, they also indicate our method to be robust under the constraints of differential privacy, unlike standard FL. Thus, \wrld is an effective approach for addressing the nascent sub-field of worldwide LM pre-training. We open several new research opportunities such as: (a) bringing the benefits of \wrld to the larger models that wealthy organizations can afford, (b) tackling broader forms of statistical heterogeneity, and (c) applying it to parameter-efficient fine-tuning. We hope this will help democratize LM training across national boundaries and address the societal concerns regarding its governance.

\begin{ack}
This work was conducted with the support of the Ministry of Education of Romania through the Credit and Scholarship Agency.
\end{ack}


{
\bibliographystyle{abbrvnat}
\bibliography{neurips}

}


\appendix
\section{Appendix}

\FloatBarrier
\begin{table}[h]
\centering
\caption{Architecture details and local training parameters for our $75$M and $250$M models. They represent the number of transformer blocks, hidden model dimension, number of attention heads, the linear layer expansion ratio and the parameters of Adam. }
\label{tab:lm_hyperaparams}
\begin{tabular}{@{}rcccccc@{}}
\toprule
    \textbf{Model (size)} &
    \textbf{\#Blocks} &
    $\boldsymbol{d}$ &
    \textbf{\#Heads} &
    \textbf{Exp.~Ratio} &
    $\mathbf{(\boldsymbol{\beta_1},~\boldsymbol{\beta_2})}$ &
    $\boldsymbol{|\textbf{Vocab}|}$ \\ \midrule
\textbf{English~($\mathbf{75}$M)} & 3 & 896 & 16 & 4  & $(0.9,~0.95)$ & $50$K \\
\textbf{English~($\mathbf{125}$M)} & 12 & 768 & 12 & 4 & $(0.9,~0.95)$ & $50$K \\
\textbf{Multi~($\mathbf{250}$M)} & 3 & 896 & 16 & 4 & $(0.9,~0.95)$ & $250$K \\ \bottomrule
\end{tabular}
\end{table}

\begin{figure}[ht]
    \centering
    \subfloat[]{\includegraphics[width=0.5\textwidth]{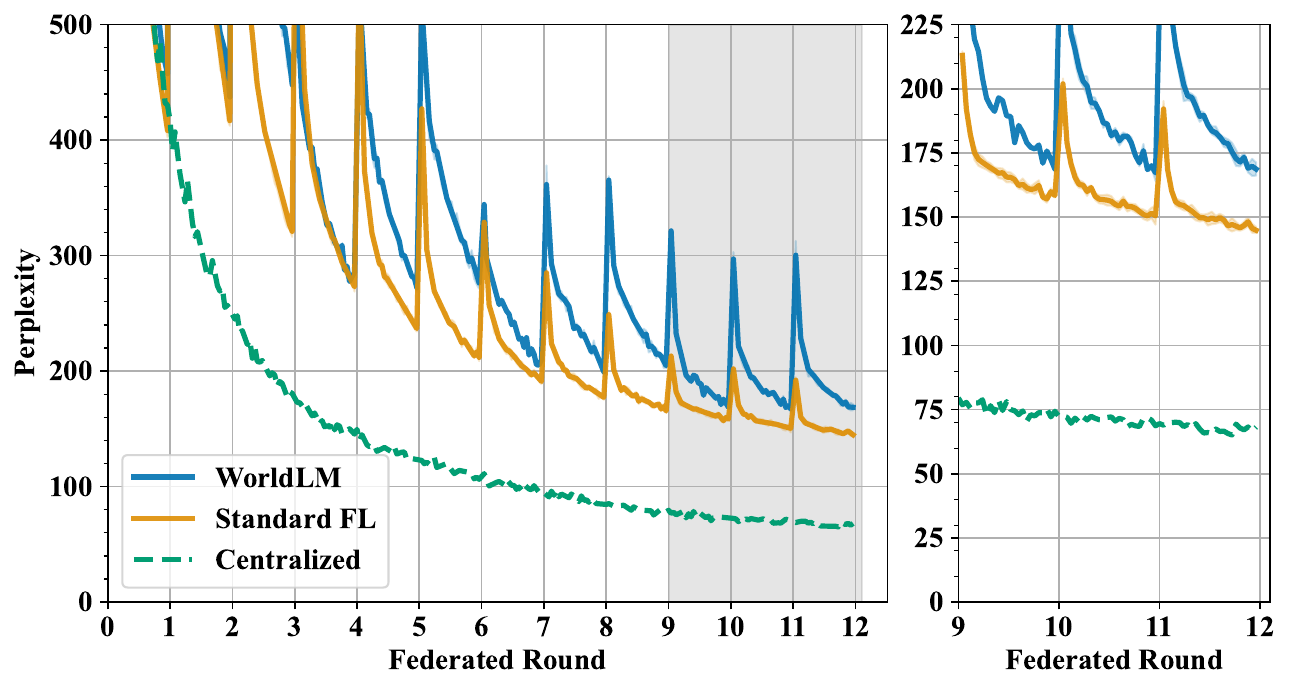}} 
    \subfloat[]{\includegraphics[width=0.5\textwidth]{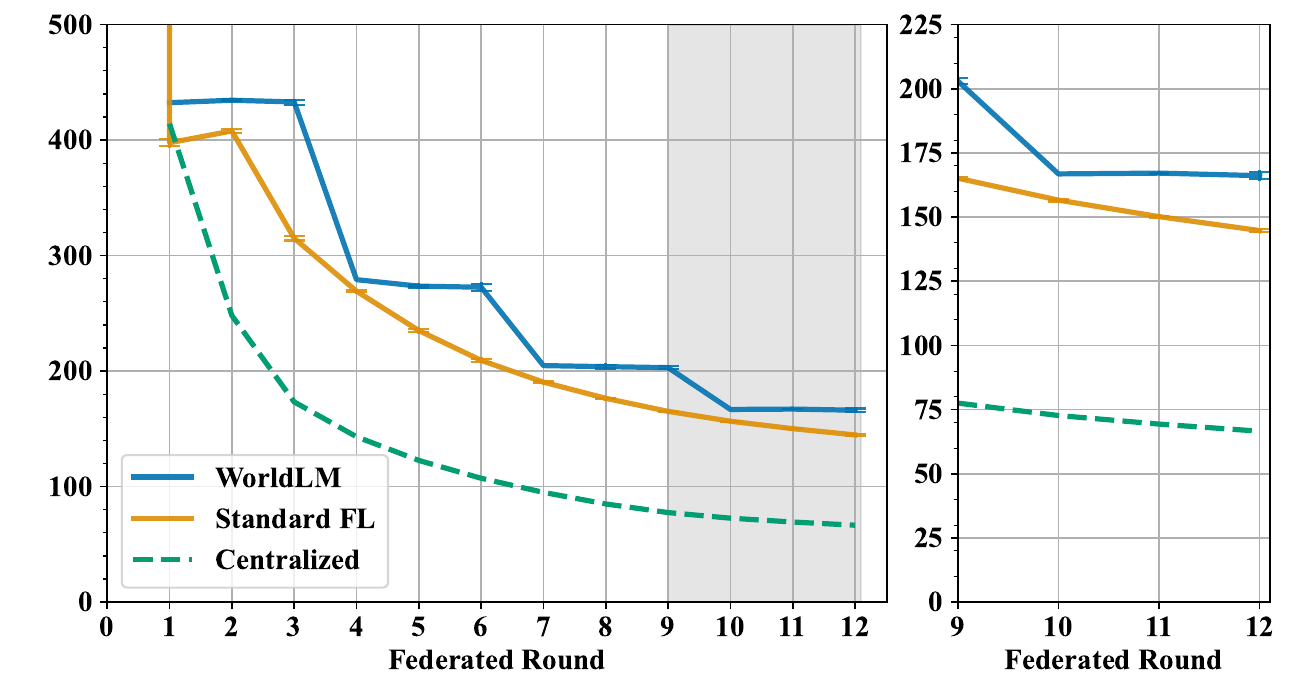}}\\
    \caption{\wrld training~(a) and validation~(b) performance of the $75$M English model on an \textit{IID} partitioning of the C4 dataset. Unlike the results for the heterogeneous partitionings of \emph{The Pile} and \emph{mC4}, \wrld underperforms standard FL as the personalization of attention-based aggregation does not provide benefits. We show a centralized model instead of purely local models, as the \textit{IID} partitioning makes any differences between them minor.}
    \label{app:fig:fed:pre_train_C4_ppl}
\end{figure}

\begin{figure}[ht]
    \centering
    \subfloat[]{\includegraphics[width=0.5\textwidth]{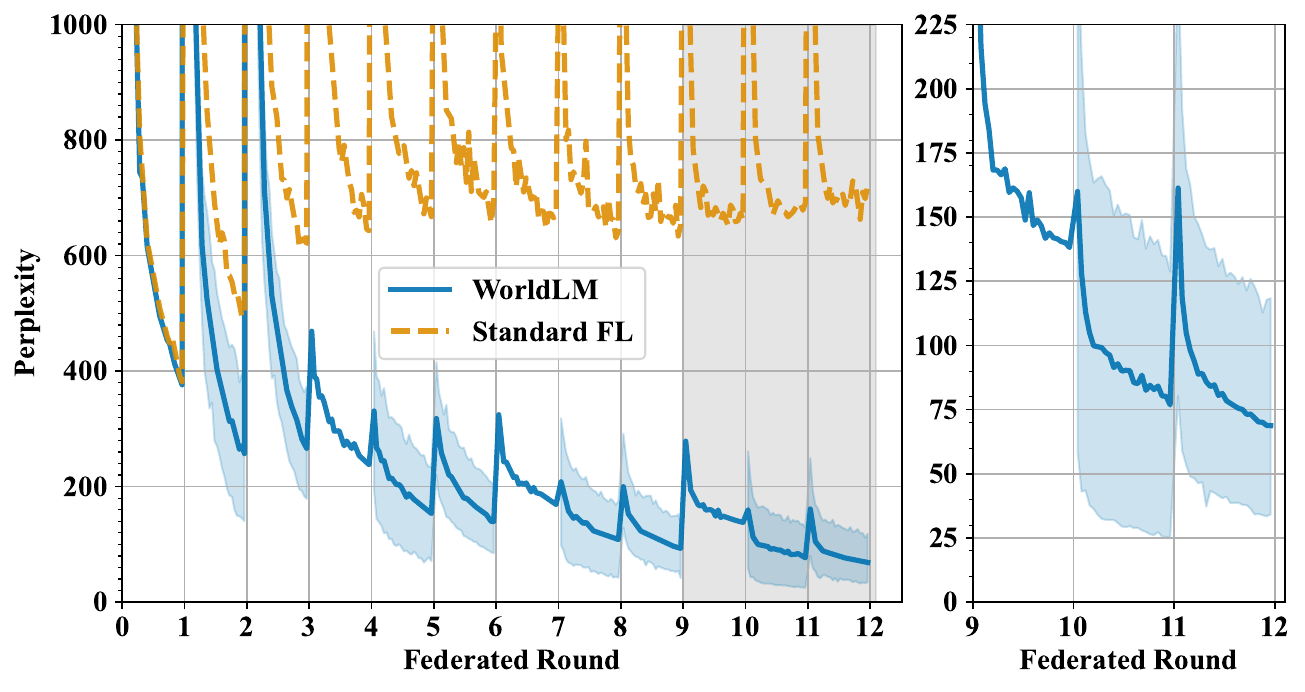}} 
    \subfloat[]{\includegraphics[width=0.5\textwidth]{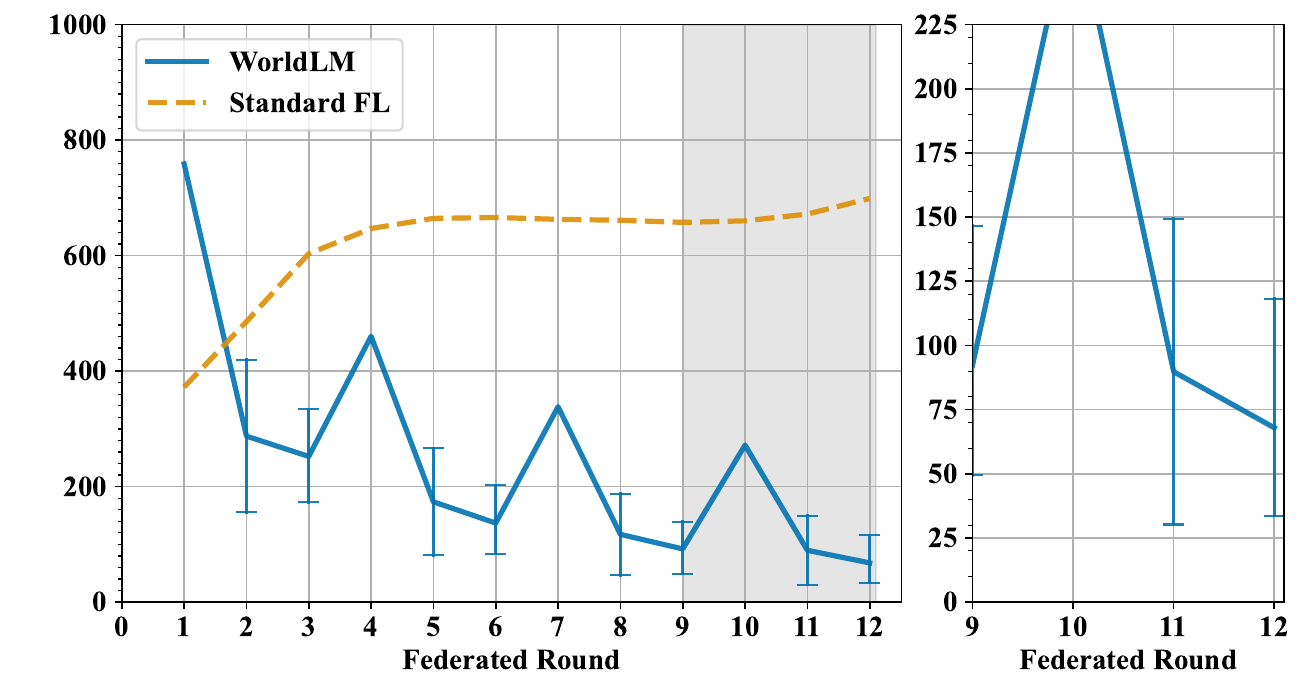}}\\
    \caption{The impact of differential privacy on \wrld performance, with the $75$M model on \emph{The Pile}, versus standard FL when using Gaussian noise with $\sigma=0.5$ and clipping client pseudo-gradients to the median pre-clip norm of the previous round. \wrld is able to stabilize the training and validation performance~(a,b) when applying DP over the leaf \textbf{PBC} and \textbf{PBA} clients, unlike FL which diverges.}
    \label{app:fig:fed:dp_23_05}
\end{figure} 

\FloatBarrier

\subsection{The Legal Context of LLM Training}\label{app:legal_context}

The surge in popularity of language models, notably exemplified by the release and widespread adoption of ChatGPT, has accelerated the integration of AI into various sectors. This expansion has subsequently encouraged the development of regulatory frameworks to govern AI technologies. A pioneering effort in this domain is the European Union's Artificial Intelligence Act (the EU AI Act), which represents the first comprehensive legal framework of its kind, anticipated to set a precedent for global AI regulation \citep{woisetschlager2024federated}. The Act encompasses, among various provisions, rigorous data governance guidelines (Art. 10), including adherence to the General Data Protection Regulation (GDPR). This introduces significant challenges for AI developers, particularly concerning the international transfer of data and data de-biasing processes.

\textit{Restrictions in cross-border transfer of data}: The EU's GDPR imposes stringent conditions on the international transfer of personal data, particularly to third countries deemed to lack adequate protection for personal data. The criterion for 'adequacy', as established in Schrems v DPC (C-262/14)) and Recital 104, requires a level of protection 'essentially equivalent' to that of the EU, a high bar for international data transfers, especially to developing countries. Additionally, the requirement for periodic reviews of adequacy for jurisdictions considered equivalent, alongside mandated safeguard measures (Art. 49) for transfers to non-equivalent third countries, introduces a layer of uncertainty and financial burden for businesses engaged in data transfer. Moreover, the EU is not the only jurisdiction tightening controls over data transfer. China, for instance, has enacted laws and supplementary provisions mandating a security assessment by regulator for the transfer of 'important data' or personal data exceeding specific thresholds, barring certain exemptions. The varied landscape of regulations poses challenges for accessing diverse local datasets while adhering to disparate, and sometimes significantly different, regulations across jurisdictions. Hierarchical FL offers an efficient solution to maintain compliance by storing data and model within its jurisdiction of origin, avoiding cross-border transfers.

\textit{Mitigation of data bias}: The EU AI Act mandates rigorous oversight throughout the entire lifecycle of the data used in AI models and obligates the implementation of \textit{'appropriate measures to detect, prevent, and mitigate potential biases'} (Art. 10.2f,fa). It underscores the growing importance of accessing diversified data sources, particularly those from jurisdictions that are underrepresented. To accommodate the acquisition of such data while adhering to individual privacy and local regulations, a novel training paradigm, comprising multiple layers of federation both among clients and jurisdictions, is necessitated to address the limitations of the traditional single-layer FL framework.

\textit{The right to information}: Transparency regarding the data collected and utilized in training, along with the rights of AI developers to access such information, has garnered considerable attention from both regulators and content creators. The GDPR grants individuals the right to obtain all information stored by a service provider (Art. 15, Rec. 63 \& 64), including details on the application of this data in training models. Furthermore, the EU AI Act mandates that model providers compile and disclose a comprehensive summary of the training content publicly (Art. 52c). In the US, legal disputes such as "Times v OpenAI" have underscored the debate over the extent to which the fair use doctrine under US copyright law protects the utilization of copyrighted materials in the training of AI models. This case also ignites broader discussions about the adequacy of the current legal framework in safeguarding content creators against the opaque practices of LLM training. These challenges have led to legislative proposals, including the Generative AI Copyright Disclosure Bill in the US House of Representatives, aimed at enhancing transparency and accountability. Hierarchical FL, by its design, offers inherent advantages over centralized training models by delineating clear data provenance — identifying the sources of data and their contributors. This attribute of Hierarchical FL positions it favorably in addressing concerns related to informational rights and data privacy, presenting a more transparent framework for data utilization in AI development.

\textit{Energy efficiency}: The EU AI Act advocates for the environmentally sustainable development of AI systems by proposing the formulation of a Code of Conduct. This Code is intended to establish explicit objectives and key performance indicators (Art. 69), mirroring the core values of the EU but also responding to growing concerns within the industry and broader society regarding the energy consumption associated with the training and use of AI. 


\newpage

\end{document}